\documentclass[11pt]{article}

\usepackage[final]{acl}

\usepackage{times}
\usepackage{latexsym}
\usepackage{tabularx}
\usepackage{booktabs}
\usepackage{amsmath}

\usepackage{graphicx}
\usepackage{listings}
\usepackage{caption}
\usepackage{xcolor} 

\usepackage[T1]{fontenc}

\usepackage[utf8]{inputenc}

\usepackage{microtype}

\usepackage{inconsolata}

\usepackage{graphicx}

\title{MasonNLP at MEDIQA-WV 2025: Multimodal Retrieval-Augmented Generation with Large Language Models for Medical VQA}

\author{A H M Rezaul Karim \\
  George Mason University, VA, USA\\
  \texttt{akarim9@gmu.edu} \\\And
  \"Ozlem Uzuner\\
  George Mason University, VA, USA\\
  \texttt{ouzuner@gmu.edu} \\}

\begin{document}
\maketitle
\begin{abstract}
Medical Visual Question Answering (MedVQA) enables natural language queries over medical images to support clinical decision-making and patient care. The MEDIQA-WV 2025 shared task addressed wound-care VQA, requiring systems to generate free-text responses and structured wound attributes from images and patient queries. We present the MasonNLP system, which employs a general-domain, instruction-tuned large language model with a retrieval-augmented generation (RAG) framework that incorporates textual and visual examples from in-domain data. This approach grounds outputs in clinically relevant exemplars, improving reasoning, schema adherence, and response quality across dBLEU, ROUGE, BERTScore, and LLM-based metrics. Our best-performing system ranked 3\textsuperscript{rd} among 19 teams and 51 submissions with an average score of 41.37\%, demonstrating that lightweight RAG with general-purpose LLMs—a minimal inference-time layer that adds a few relevant exemplars via simple indexing and fusion, with no extra training or complex re-ranking— provides a simple and effective baseline for multimodal clinical NLP tasks. \footnote{Implementation can be found here: \url{https://github.com/AHMRezaul/MEDIQA-WV-2025}}
\end{abstract}

\section{Introduction}

Generating accurate answers to clinically relevant questions about medical images, known as Medical Visual Question Answering (MedVQA), requires integrating visual perception with domain-specific reasoning \citep{lin2023medical, lau2018dataset}.
 Such systems can enhance diagnostics, support clinical training, and provide accessible, question-driven insights for clinicians and patients.

Compared to general VQA, MedVQA faces unique challenges, such as subtle anatomical or pathological features that must be interpreted precisely, and questions often demanding specialized knowledge and logical inference \citep{lin2023medical, liu2021slake}. General VQA datasets lack this depth, motivating the creation of tailored medical benchmarks \citep{lin2023medical}. Key resources include VQA-RAD for radiology \citep{lau2018dataset}, SLAKE with bilingual semantic annotations \citep{liu2021slake}, and ImageCLEF’s VQA-Med series \citep{ben2019vqa, ben2021overview}. PathVQA extends to pathology images \citep{he2020pathvqa}, PMC-VQA scales to over 227k Question Answer pairs for pretraining \citep{zhang2023pmc}, and MedFrameQA introduces multi-image reasoning for clinical scenarios \citep{yu2025medframeqa}. While these datasets drive progress, many methods still rely on resource-intensive fine-tuning and large domain corpora, limiting scalability.

Wound-care is a crucial MedVQA application, where image-based assessment guides treatment, monitors healing, and detects complications. Remote wound monitoring and telemedicine reduce costs, hospital visits, and infection risks \citep{sood2016role, chen2020telemedicine}, but variability in interpretation highlights the need for automated QA tools to support clinicians and empower patients.

\textbf{The MEDIQA-WV shared task} (Wound-care Visual Question Answering), part of ClinicalNLP 2025, addresses this challenge by generating free-text answers to patient-oriented wound-care questions using one or more images with annotations \citep{MEDIQA-WV-Task}. The shared task dataset includes bilingual (English/Chinese) queries, metadata such as wound type and anatomic site, and systems are evaluated on fluency, relevance, and clinical accuracy.

We study an instruction-tuned general-domain LLM \texttt{(Meta LLaMA-4 Scout 17B)} \cite{meta2025llama} in a few-shot setup. It performs well on cases with small image details and short, generic question types, but degrades on images with subtle or mixed findings, multi-part questions, and requests that require expert-level interpretation. To improve grounding and reasoning, we add a lightweight retrieval-augmented generation (RAG) \cite{lewis2020retrieval} layer by retrieving top-2 relevant text and image exemplars from the task corpus and appending them to the prompt. Since the dataset is not large enough for reliable fine-tuning and would add substantial compute and operational cost, a lightweight RAG setup was chosen.

\begin{figure*}[ht]
\centering
\includegraphics[width=\textwidth]{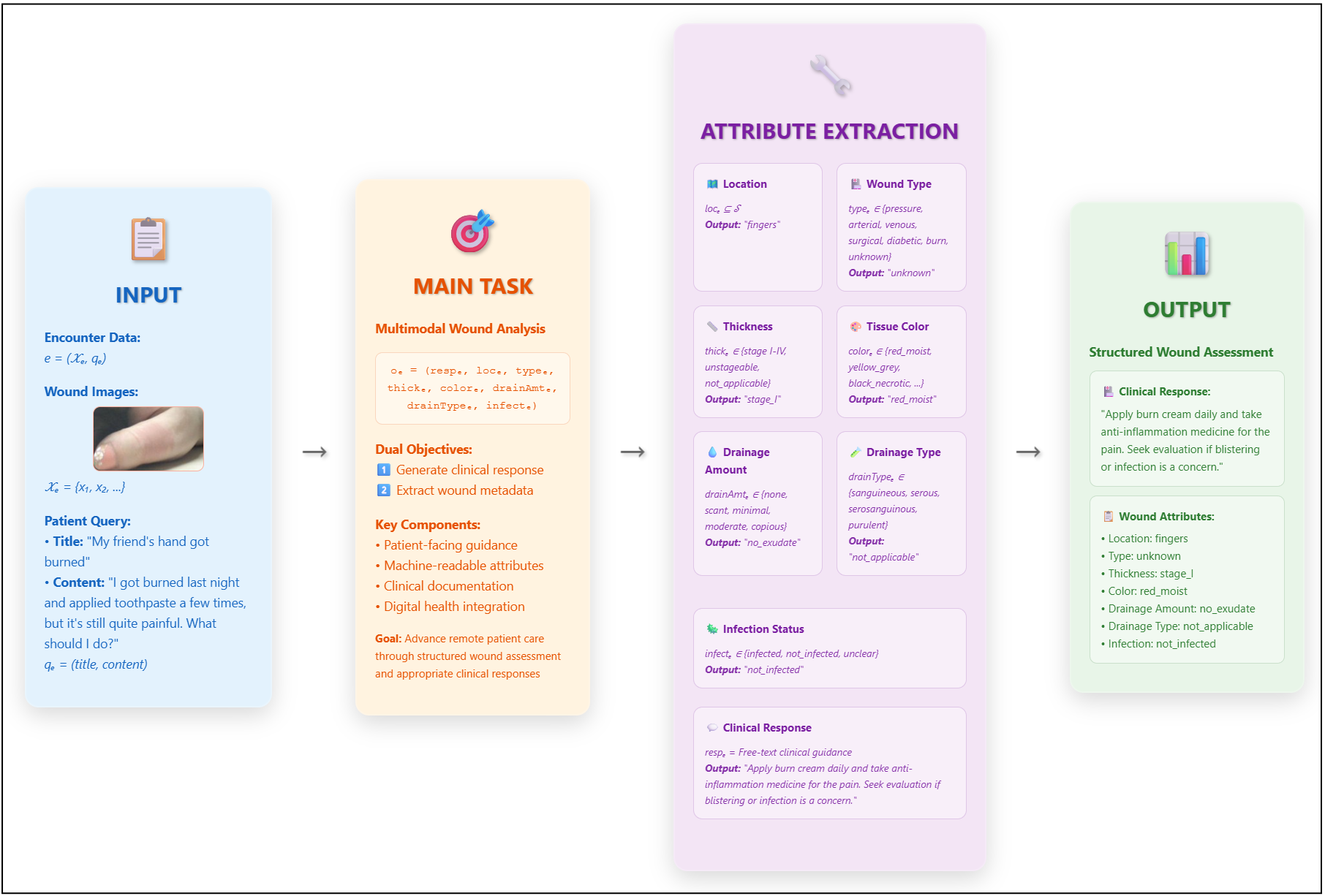} 
\caption{Task overview for MEDIQA-WV 2025. Inputs: wound images and a patient query. Outputs: free-text answer with structured wound attributes}
\label{fig:task}
\end{figure*}

Our contributions include:
\begin{itemize}
\item Demonstrating that a general-domain LLM with lightweight RAG can handle complex multimodal clinical tasks without domain-specific training.
\item Showing that exemplar retrieval at inference improves reasoning and interpretability on clinical data.
\item Providing a systematic analysis of how retrieval modality (text-only vs. multimodal) and prompting choices affect performance in medical visual question answering.
\end{itemize}

These results illustrate the promise of general-purpose LLMs, augmented with lightweight RAG, for transparent, flexible, and efficient solutions in clinical NLP and multimodal AI.

\section{Related Work}

Early VQA systems in both general and clinical domains relied on rule-based pipelines and small answer vocabularies, mapping hand-crafted cues or shallow features to fixed slots. These approaches lacked robustness to negation, uncertainty, and paraphrase \cite{malinowski2015ask}. In the general domain, although VQA was framed as open-ended, many methods treated it as classification over restricted answer sets \citep{antol2015vqa}. Similar patterns appeared in early medical benchmarks, where evaluation emphasized exact match or lexical overlap, reinforcing closed-set, short-answer formats \citep{hasan2018overview,ben2019vqa,ben2021overview}. Such formulations constrained clinical expressivity and hindered nuanced responses.

With deep learning, convolutional image encoders combined with recurrent or simple text encoders became standard, later enhanced by attention \cite{talafha2018just,lin2023medical}. In the general domain, bottom-up/top-down attention over regions \citep{anderson2017bottom} and modular co-attention \citep{yu2019deep} set strong baselines, influencing medical adaptations \citep{lin2023medical}. New datasets supported this shift: VQA-RAD \citep{lau2018dataset} introduced clinically authored radiology questions; SLAKE \citep{liu2021slake} added bilingual annotations with semantic labels; PathVQA \citep{he2020pathvqa} scaled pathology QA with textbook images but faced noise and coverage issues; Medical-Diff-VQA \citep{hu2023medical} introduced difference-based paired-image questions for comparative reasoning.

Transformer-based vision–language pretraining further reshaped the field. ViLBERT \cite{lu2019vilbert} and LXMERT \citep{tan2019lxmert} learned joint cross-modal representations and adapted effectively to VQA. In medicine, MMBERT \citep{khare2021mmbert} showed multimodal BERT \cite{devlin2019bert} pretraining improves MedVQA under data scarcity, and M2I2 \citep{li2023self} leveraged self-supervised masked modeling and contrastive alignment to advance results across VQA-RAD, PathVQA, and SLAKE. Hybrids also emerged: BPI-MVQA \citep{liu2022bpi} combined transformers with retrieval signals for improved multimodal fusion. These approaches improved accuracy but generally required domain-specific pretraining or fine-tuning.

Large vision–language models (VLMs) and LLM–vision hybrids enabled open-ended generation. BLIP-2 \citep{li2023blip} efficiently bridged frozen encoders and LLMs. LLaVA \citep{liu2023visual} introduced visual instruction tuning, while LLaVA-Med \citep{li2023llava} adapted this strategy to biomedical content. Domain-specific conversational VLMs such as XrayGPT \citep{thawakar2024xraygpt} aligned MedCLIP \cite{wang2022medclip} encoders with Vicuna \cite{chiang2023vicuna} for chest X-ray QA and summarization, and R-LLaVA \citep{chen2024r} enhanced MedVQA via ROI annotations. Generative perspectives also gained traction: PMC-VQA scaled to 227k QA pairs, training MedVInT for effective fine-tuning on VQA-RAD, SLAKE, and ImageCLEF \citep{zhang2024development}. Evaluation evolved from strict accuracy toward BLEU and other text-generation metrics to capture partial correctness and phrasing variability \citep{ben2019vqa,ben2021overview,hasan2018overview}.

RAG \cite{lewis2020retrieval} has emerged to mitigate hallucinations and data scarcity by grounding answers in evidence. RAMM \citep{yuan2023ramm} combined retrieval with dedicated attention modules to set state-of-the-art results on multiple MedVQA datasets. Fine-grained retrieval fusion with re-weighting further improved benchmarks like PathVQA and VQA-RAD without direct data access \citep{liang2025fine}. Broader studies show retrieval strategies, granularity, and fusion strongly affect factuality, though best practices remain unsettled \citep{xiong-etal-2024-benchmarking}. 

Despite progress, challenges remain. Many systems rely on costly pretraining, curated corpora, or complex fusion stacks that limit transferability. Closed-set classification constrains answer diversity, while generative models risk hallucination if ungrounded. Our work addresses these issues with a general-domain, instruction-tuned LLM and lightweight RAG, which is a minimal, inference-time retrieval layer that adds a few relevant snippets via simple indexing and fusion, without extra training or complex re-ranking, to reduce hallucinations, respect data limits, and keep the system easy to reproduce. This approach of LLMs with RAG-based textual and visual exemplars preserves generative flexibility while improving interpretability and reproducibility by grounding answers in retrieved evidence, aligning with pragmatic, evidence-grounded MedVQA.

\section{Task Description}

The MEDIQA-WV shared task \cite{MEDIQA-WV-Task} extends prior efforts in MedVQA to the wound-care domain. The objective is to advance remote patient care by generating clinically appropriate free-text responses to patient queries, while at the same time producing structured wound-related metadata that capture essential clinical details. This dual requirement reflects the need for both patient-facing guidance and machine-readable data that can be integrated into electronic health records (EHR).

Formally, each data instance corresponds to an \emph{encounter} $e$, defined as a pair $(\mathcal{X}_e, q_e)$. The image set $\mathcal{X}_e = \{x^{(1)}_e,\dots,x^{(n)}_e\}$ contains one or more wound photographs, and the textual query $q_e$ is bilingual, consisting of an English and a Chinese title and content.

The system must predict an output tuple with a response and the following metadata.
\[
o_e = (\textit{resp}_e, \textit{loc}_e, \textit{type}_e, \textit{thick}_e,\]\[ \textit{color}_e, \textit{drainAmt}_e, \textit{drainType}_e, \textit{infect}_e),
\]
Where $\textit{resp}_e$ is a free-text response and the remaining fields represent structured wound metadata. The anatomic location $\textit{loc}_e \subseteq \mathcal{L}$ may include one or more sites (e.g., arm, chest, foot). The wound type $\textit{type}_e \in$ \{ \textit{pressure, arterial, venous, surgical, diabetic, ...} \} covers common etiologies. The wound thickness $\textit{thick}_e \in \{ \textit{stage I–IV, unstageable, not\_applicable} \}$. The tissue color $\textit{color}_e$ is drawn from a finite set describing visual appearance (e.g., \textit{red/moist, yellow/grey, black/necrotic}). Drainage is captured both in amount, $\textit{drainAmt}_e \in \{ \textit{none, scant, minimal, moderate, copious} \}$, and in type, $\textit{drainType}_e \in$ \{ \textit{sanguineous, serous, serosanguinous, purulent}\}. Finally, the infection status $\textit{infect}_e \in \{\textit{infected}, \textit{not\_infected}, \textit{unclear}\}$.

Training data provide full tuples $o_e$ for each encounter, while in the test phase, only $(\mathcal{X}_e, q_e)$ are given and systems must predict $\hat{o}_e$. Success in this task requires models to jointly reason over multimodal inputs, differentiate clinically meaningful features, and generate outputs that are both fluent and structured for downstream clinical use.

\section{Dataset}

The MEDIQA-WV dataset \cite{MEDIQA-WV-2025-WoundcareVQA-Dataset} was created to support wound assessment and patient counseling tasks. Each encounter consists of a unique identifier, one or more wound images, a bilingual query in English and Chinese, and a set of expert-generated responses in both languages. In addition to the free-text components, the training and validation splits contain structured gold-standard metadata covering the following attributes: \texttt{wound\_type, wound\_thickness, tissue\_color, drainage\_amount, drainage\_type, infection\_status,} and one or more \texttt{anatomic\_locations}. All categorical values are drawn from a closed dictionary of medically valid terms, such as wound types \{\textit{traumatic, surgical, pressure}\}, tissue colors \{\textit{red moist, necrotic black}\}, drainage categories specifying both \textbf{amount} and \textbf{type}, and anatomic sites like \textit{arm, knee, foot}. Figure \ref{fig:task} demonstrates an example data instance.

\begin{table}[h]
\centering
\begin{tabular}{lccc}
\toprule
\textbf{Split} & \textbf{Encounters} & \textbf{Responses} & \textbf{Images} \\
\midrule
Train & 279 & 279 & 449 \\
Validation & 105 & 210 & 147 \\
Test & 93 & 279 & 152 \\
\bottomrule
\end{tabular}
\caption{Dataset statistics: encounters, responses, and images per split.}
\label{tab:dataset-splits}
\end{table}

\subsection{Dataset Analysis}

Table~\ref{tab:dataset-splits} summarizes the distribution of encounters, responses, and images across splits. The training set provides a single expert response per encounter, while validation is double-annotated, offering complementary perspectives. The test set is input-only and triple-annotated by medical professionals, though the gold-standard labels remain unpublished.

Encounters contain varying numbers of images, reflecting the clinical setting where multiple photos capture different wound angles or progress. In the training split, 170 encounters include a single image, while 109 (39\%) contain multiple (up to nine) images. Validation includes 72 encounters with single images and 33 encounters with multiple images, and the test set has 55 single-image and 38 multiple-image encounters. Both the validation and test sets contain up to four images for a single encounter. Queries and responses also differ across splits. English queries average 46 words in training, 44 in validation, and 52 in test. Responses are 29 words on average for training, but become longer in validation (41 words) and test (47 words). 

The metadata distribution is highly skewed. Traumatic wounds dominate with 330 cases (85.9\%), while arterial and venous ulcers appear only once each (0.3\%). Infection status is similarly imbalanced: 325 encounters (84.6\%) are labeled as not infected, 39 as unclear (10.2\%), and only 20 as infected (5.2\%). Wound thickness is concentrated in stage~I and stage~II, and common anatomical sites include the lower leg, fingers, and hand. Although annotations generally follow the predefined dictionary, occasional inconsistencies appear, such as \textit{“sole”} instead of \textit{“foot-sole”} or drainage mismatches like \textit{“no exudate”} paired with a specific drainage type. These rare cases highlight the need for normalization.

Overall, the dataset integrates structured wound metadata, bilingual queries, and expert responses into a challenging benchmark. The skewed label distributions and queries with multiple images, and the small size of training data, make fine-tuning difficult. These properties motivate using an LLM with RAG to retrieve similar examples from the training data, so answers stay close to the data, avoid generic responses, and follow the required output format.

\begin{figure}[ht]
\centering
\includegraphics[width=\columnwidth]{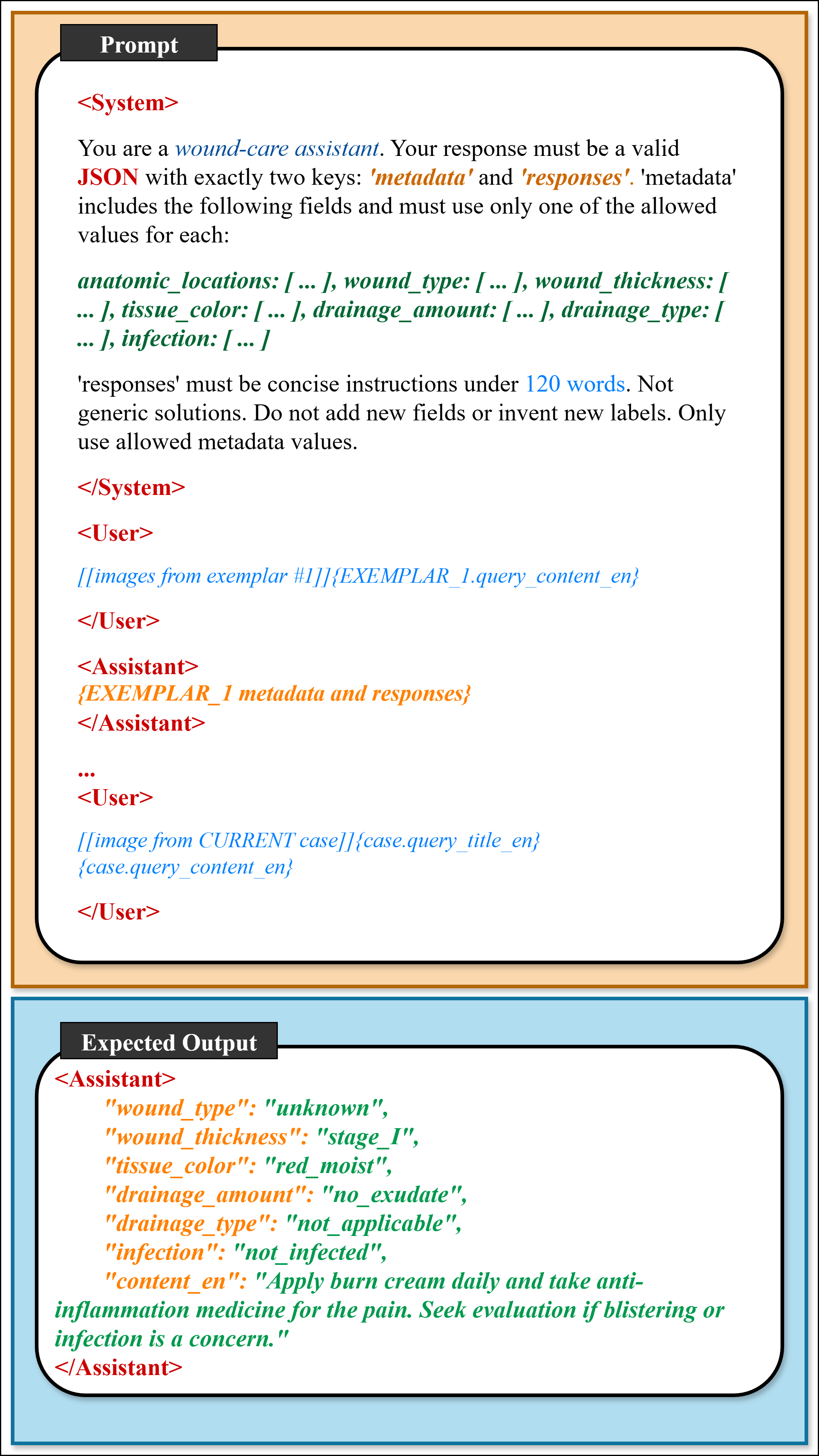} 
\caption{Structured prompt with retrieved exemplars and the expected output schema.}
\label{fig:prompt}
\end{figure}

\section{Methodology}
To test how a general-domain LLM performs on a MedVQA task without domain-specific training, the \textit{meta-llama/Llama-4-Scout-17B-16E-Instruct} \cite{meta2025llama} model was chosen. It follows instructions well, has open weights for reproducible research, and is a strong multimodal variant in the Meta-LLaMA \cite{touvron2023llama} family, offering a long context window and reliable vision-language support.
\subsection{Model Configuration}
We used the 17B instruction-tuned LLaMA-4 model, implemented via Hugging Face \texttt{transformers} with automatic GPU mapping. Inference ran in \texttt{bfloat16} for efficiency, with a maximum generation length of 4096 tokens, temperature 0.2, and top-p 0.9. For multimodal inputs, the model was paired with the LLaMA-4 processor to jointly encode text prompts and wound images.

\subsection{Prompt Design}
We explored three prompting strategies: zero-shot, few-shot, and RAG. An example prompt is provided in Figure~\ref{fig:prompt}.

\paragraph{Zero-shot prompting.}
The model received only a system instruction defining its role as a wound-care assistant. Outputs were constrained to valid JSON by dividing the output tuple into two top-level keys: \texttt{metadata} and \texttt{responses}. \texttt{Metadata} used categorical labels from a wound-care data dictionary (e.g., wound type, tissue color, drainage, infection status), while \texttt{responses} provided short patient-facing instructions ($\leq$120 words). This setting tested schema adherence without exemplars.

\paragraph{Few-shot prompting.}
We added two exemplar encounters from the training set, chosen after evaluating on the validation set, to reduce schema violations and improve metadata consistency. Each exemplar included wound image(s) and query text as a user turn, followed by the reference response as an assistant turn, guiding the model to emulate JSON structure and style. We limit exemplars to two because adding more, together with images, metadata, and the current prompt, exceeds the model’s context window.

\paragraph{Retrieval-augmented prompting.}
To improve grounding and reduce hallucinations \cite{lewis2020retrieval}, we designed a multimodal RAG pipeline combining dense similarity search with exemplar-driven prompting, where we encoded questions and images into vectors, then retrieved the nearest training examples for that encounter and placed those exemplars in the prompt. Two indices were built with FAISS \cite{douze2024faiss}: semantic text embeddings from \texttt{sentence-transformers/all-MiniLM-L6-v2} and vision–language embeddings from CLIP (\texttt{openai/clip-vit-base-patch32}) \footnote{\href{https://huggingface.co/sentence-transformers/all-MiniLM-L6-v2}{sentence-transformers}, \href{https://huggingface.co/openai/clip-vit-base-patch32}{openai-clip}}.
We tested both the text-only and multimodal (text+image) retrieval setup.


At inference, we retrieve training encounters most similar to the inference-case using combined text and image similarity with equal weight (\(\alpha = 0.5\)).
We evaluated other \(\alpha\) values that placed more weight on images, but performance declined with more weight for the image, and so an approach with image-only retrieval was not explored. We select the top two exemplars because validation runs gave the best overall metrics, and adding more with images and metadata caused the prompt to exceed the model's context window.
This setup reduced schema violations, improved metadata predictions, and outperformed zero- / few-shot prompting.

\subsection{Experimental Setup}
Images were resized to $224 \times 224$ and passed with text. Decoding used nucleus sampling without beam search to balance diversity and format compliance. All runs were performed on NVIDIA A100 GPUs (80 GB), enabling full 17B model inference with multimodal inputs. We logged raw generations to audit both successful and erroneous outputs.

\begin{table*}[ht]
\centering
\scriptsize
\resizebox{\textwidth}{!}{
\begin{tabular}{lccccccccccc}
\toprule
\textbf{Team} & \textbf{dBLEU} & \textbf{R1} & \textbf{R2} & \textbf{RL} & \textbf{RLsum} & \textbf{BERT-mn} & \textbf{BERT-mx} & \textbf{DeepSeekV3} & \textbf{Gemini} & \textbf{GPT-4o} & \textbf{Avg} \\
\midrule
\textbf{MasonNLP} & \textbf{8.89} & 70.99 & \textbf{48.62} & 42.19 & 42.27 & 59.01 & 63.27 & 53.55 & 55.38 & \textbf{55.38} & \textbf{41.37} \\
MasonNLP & 7.31 & \textbf{72.79} & 48.44 & \textbf{43.31} & \textbf{43.25} & \textbf{60.42} & \textbf{64.55} & \textbf{58.92} & \textbf{56.45} & 53.23 & 41.07 \\
\hline
EXL Services–Health & 9.92 & \textit{79.09} & \textit{56.13} & \textit{45.61} & 45.60 & \textit{62.18} & 66.90 & \textit{68.23} & \textit{64.52} & \textit{71.51} & \textit{47.30} \\
EXL Services–Health & \textit{13.04} & 71.18 & 51.28 & 45.17 & \textit{45.72} & 61.88 & \textit{67.43} & 63.49 & 59.14 & 62.90 & 45.75 \\
DermaVQA & 7.65 & 78.99 & 53.91 & 45.49 & 45.48 & 60.62 & 63.68 & 42.74 & 45.70 & 37.10 & 37.71 \\
\bottomrule
\end{tabular}}
\caption{Leaderboard results on MEDIQA-WV 2025. MasonNLP best runs in bold; best per column in italics.}
\label{tab:leaderboard}
\end{table*}
\begin{table*}[ht]
\centering
\scriptsize
\resizebox{\textwidth}{!}{
\begin{tabular}{lccccccccccc}
\toprule
\textbf{System} & \textbf{dBLEU} & \textbf{R1} & \textbf{R2} & \textbf{RL} & \textbf{RLsum} & \textbf{BERT-mn} & \textbf{BERT-mx} & \textbf{DeepSeekV3} & \textbf{Gemini} & \textbf{GPT-4o} & \textbf{Avg} \\
\midrule
\textbf{LLaMA-4 + RAG (image+text)} & \textbf{8.89} & 70.99 & \textbf{48.62} & 42.19 & 42.27 & 59.01 & 63.27 & 53.55 & 55.38 & \textbf{55.38} & \textbf{41.37} \\
LLaMA-4 + RAG (text only) & 7.31 & \textbf{72.79} & 48.44 & \textbf{43.31} & \textbf{43.25} & \textbf{60.42} & \textbf{64.55} & \textbf{58.92} & \textbf{56.45} & 53.23 & 41.07 \\
LLaMA-4 (few-shot) & 4.67 & 41.50 & 27.30 & 23.50 & 24.10 & 41.60 & 44.20 & 35.00 & 33.90 & 33.90 & 23.63 \\
LLaMA-4 (zero-shot) & 1.73 & 25.00 & 17.00 & 14.00 & 14.50 & 29.00 & 30.00 & 20.00 & 21.60 & 21.60 & 14.10 \\
\bottomrule
\end{tabular}}
\caption{Ablation of prompting and retrieval strategies. Best per column in bold.}
\label{tab:ablation}
\end{table*}

\subsection{Post-processing}
LLMs often generate extraneous text or malformed JSON, so we implemented a normalization pipeline. We first stripped any \texttt{Markdown} code fences or leading text before the opening brace, then parsed outputs to enforce exactly two keys: \texttt{metadata} and \texttt{responses}. \texttt{Metadata} entries were validated against the wound-care dictionary, discarding invalid fields. \texttt{Responses} were mapped to the English patient instruction. The cleaned output was merged into each case under its \texttt{encounter\_id}, producing the final structured predictions for evaluation.

This layered design enabled systematic comparison of zero-shot, few-shot, and retrieval-augmented prompting, quantifying the benefits of contextual grounding and exemplar retrieval on schema adherence, metadata accuracy, and response validity.

\section{Evaluation}
The MEDIQA-WV 2025 shared task employs a multi-dimensional evaluation protocol that combines surface overlap, semantic similarity, and clinical plausibility.

For lexical similarity, the task uses \texttt{deltaBLEU} \cite{galley2015deltableu}, which extends \texttt{BLEU} \cite{papineni2002bleu} by rewarding partial matches across multiple references. Complementary recall-oriented measures include \texttt{ROUGE-1}, \texttt{ROUGE-2}, \texttt{ROUGE-L}, and \texttt{ROUGE-Lsum} \cite{lin2004rouge}, capturing different levels of $n$-gram and sequence overlap.

Semantic similarity is evaluated with \texttt{BERTScore} \cite{zhang2019bertscore}, using two variants: \texttt{BERT-mn}, which averages over references, and \texttt{BERT-mx}, which takes the maximum score to reward alignment with at least one gold annotation. English responses are scored with \texttt{microsoft/deberta-xlarge-mnli} \cite{he2020deberta}, while Chinese responses are scored with \texttt{lang=zh} for multilingual alignment.

To assess plausibility and instructional quality beyond surface metrics, three large multimodal language models (LMLMs) act as automatic judges: (i) \texttt{DeepSeek-V3-0324} (Azure AI Foundry), (ii) \texttt{Gemini-1.5-pro-002} (Google GenAI), and (iii) \texttt{GPT-4o} (Azure AI Foundry)\footnote{\href{https://huggingface.co/deepseek-ai/DeepSeek-V3-0324}{DeepSeek}, \href{https://cloud.google.com/vertex-ai/generative-ai/docs/models/gemini/1-5-pro}{Gemini-1.5-pro}, \href{https://openai.com/index/hello-gpt-4o/}{GPT-4o}}. Using standardized prompts in English and Chinese, these models independently score outputs for usefulness, contextuality, and clinical appropriateness, reducing model-specific bias.

A final \texttt{average\_score (Avg)} aggregates results across all metrics, combining fidelity, semantic alignment, and plausibility into a robust benchmark for multimodal clinical generation systems.

\section{Results and Discussion}

\subsection{Leaderboard Performance}
The MEDIQA-WV 2025 shared task attracted participation from \textbf{19} teams, producing a total of \textbf{51} submissions. Our \textbf{MasonNLP} system ranked competitively, achieving an average score of \textbf{41.37\%} on its best run. As shown in Table~\ref{tab:leaderboard}, both of our submissions placed in the top five overall, underscoring the robustness of our general-domain LLM pipeline against more specialized approaches. Notably, while the leading system achieved the highest overall performance (\textit{47.30\%}), our systems demonstrated comparable strength across multiple metrics, reflecting effective phrasing and semantic alignment. This suggests that our lightweight retrieval and prompting strategies can yield results close to top-level systems.

\subsection{Ablation Study}
To better understand the contribution of the retrieval and prompting strategy, we conducted an ablation across four configurations: (1) \texttt{LLaMA-4 + RAG} with \emph{image+text} retrieval, (2) \texttt{LLaMA-4 + RAG} with \emph{text-only} retrieval, (3) \texttt{LLaMA-4} in \emph{few-shot}, and (4) \texttt{LLaMA-4} in \emph{zero-shot}. Results in Table~\ref{tab:ablation} demonstrate three key effects. First, retrieval markedly improves all evaluation metrics, confirming its role in grounding predictions. Second, the inclusion of images supplied visual evidence for image-dependent details, as shown by higher \texttt{dBLEU} and \texttt{GPT-4o} scores. Third, even without retrieval, moving from zero-shot to few-shot reduces hallucinations and yields more consistent phrasing, though the gap to retrieval-based models remains large. Together, these trends highlight that retrieval complements prompting and that multimodal retrieval is particularly effective for wound-specific guidance. This systematic progression from zero-shot to multimodal RAG reveals clear patterns in how different retrieval modalities and prompting approaches affect MedVQA performance.

\subsection{Discussion and Implications}
Our results show a clear progression in performance from zero-shot prompting to multimodal RAG. In the \emph{zero-shot} setting with the \textbf{LLaMA-4 17B} model, scores were very low (\texttt{dBLEU} 1.73), largely due to the model’s failure to produce the required structured JSON output despite explicit instructions. 

Adding a few in-context exemplars improved formatting and raised \texttt{dBLEU} to 4.67, but responses remained generic and lacked clinically specific detail. Retrieval with \emph{textual exemplars} addressed this issue more effectively. By grounding outputs in semantically similar queries and solutions, the model produced more structured and concrete recommendations, with \texttt{Rouge-L} increasing from 23.50 (few-shot) to 43.31, and \texttt{GPT-4o} judgments rising substantially. 

Extending retrieval to include \emph{images} further boosted contextual grounding, particularly for wound-site descriptions and infection cues, lifting \texttt{dBLEU} to 8.89. However, gains were not universal. Visual neighbors sometimes introduced noise when image relevance was weak, slightly trailing text-only retrieval in a few metrics. 

Overall, the ablation confirms that moving from zero-shot to exemplar-based and multimodal retrieval progressively improves structure and specificity. A lightweight RAG pipeline combining textual and visual evidence provides a strong, reproducible baseline for multimodal clinical tasks without domain-specific fine-tuning.

\begin{table*}[t]
\centering
\small
\setlength{\tabcolsep}{6pt}
\renewcommand{\arraystretch}{1.5}
\begin{tabularx}{\textwidth}{>{\raggedright\arraybackslash}p{0.20\textwidth}|>{\raggedright\arraybackslash}X|>{\raggedright\arraybackslash}X}
\hline
\textbf{Improvement Type} & \textbf{Zeroshot Prediction} & \textbf{RAG Prediction} \\
\hline
Hallucination Reduction &
\textit{Infection: infected} \newline
Instruction: ``Antibiotics may be needed.'' &
\textit{Infection: not\_infected} \newline
Instruction: ``No signs of infection; continue saline cleaning and dry dressing.'' \\
\hline
Specificity of Response &
\textit{Location: finger} \newline
Instruction: ``Keep the area clean and avoid movement.'' &
\textit{Location: fingertip} \newline
Instruction: ``Clean fingertip wound twice daily, apply antibiotic ointment, and avoid immersion in water.'' \\
\hline
Vocabulary Normalization &
\textit{Type: trauma} \newline
Instruction: ``Healing depends on care.'' &
\textit{Type: traumatic} \newline
Instruction: ``Traumatic wound; healing time approx.\ 2--3 weeks with proper care.'' \\
\hline
\end{tabularx}
\caption{Examples of improvements from zero-shot to RAG, grouped by improvement type.}
\label{tab:rag_improvements}
\end{table*}

\section{Error Analysis}
In the absence of gold-standard labels, we evaluate model behavior along four axes: (i) schema conformance against an allowed-value dictionary, (ii) content form and genericness (length, template reuse, lexical alignment to the query), (iii) intent coverage for common asks (healing time, stitches/ sutures, tetanus), and (iv) hallucination/ over-claim heuristics (e.g., asserting infection without cues).

\subsection{Zero-shot \textsc{LLaMA-4}}
On 93 queries, the model produced 93 answers with one empty reply (1.1\%). Answers are short (mean 18.1 words with max 53) and frequently reuse stock advice, like ``cover with a bandage'' (25/93), ``monitor for signs of infection'' (23/93), ``apply antibiotic ointment'' (22/93), with additional phrases such as ``seek medical attention'' (9/93), ``consult a doctor'' (6/93), and ``keep the area clean and dry'' (5/93). Although 90 outputs are unique (only two duplicates and one missing), query–answer lexical overlap is low, indicating a generic style that often under-engages the user’s ask. Intent coverage lacks precision as well. For healing-time questions, only 1/16 answers include a numeric time frame; for stitches/ sutures, 4/13 mention suture care or removal timing; for tetanus, 4/7 mention vaccination/ booster guidance. Hallucination screening flags 31/93 answers that assert infection without any infection cues in the corresponding queries; about a quarter of these are hedged (e.g., ``\textit{may be} infected''), and explicit speculative diagnosis terms (e.g., fracture, necrosis) are rare (4/93). Overall, zero-shot outputs are fluent and safety-oriented but frequently generic, under-answer explicit asks, and sometimes over-call infection in the absence of evidence.

\subsection{\textsc{LLaMA-4} + RAG (Image+Text)}
We examined 93 predictions for schema conformance, value validity, and content quality. All seven fields were present for every item. True out-of-vocabulary (OOV) rates were low as \emph{anatomic\_locations} had 8 OOV entries driven by common synonyms (\textit{leg}, \textit{finger/fingertip}, \textit{shin}), while single-valued fields each had at most one OOV instance (\emph{wound\_type} 1/93; \emph{wound\_thickness} 4/93 due to \textit{partial}/\textit{partial thickness}; \emph{tissue\_color}, \emph{drainage\_amount}, \emph{drainage\_type}, \emph{infection} each 1/93). Label distributions reflected the training and development set analysis with \emph{wound\_type} mostly being \textit{traumatic} (88.0\%), \emph{infection} favoring \textit{not\_infected} (52.2\%) with mass on \textit{infected} (27.2\%) and \textit{unclear} (20.7\%), and \emph{wound\_thickness} was dominated by \textit{stage\_II} (50.6\%). There was exactly one instance with no generated response. Responses were longer than the zero-shot system (mean 28.4 words with a max of 96) and remained largely unique (91/93) but still exhibited a generic tone. About 60\% of answers had very low lexical overlap with their queries, and common advice tokens were frequent (e.g., ``antibiotic'' in 45.2\%; ``debridement'' in 5.4\%). Intent coverage improved but remained uneven. 7/44 (15.9\%) healing-time questions received a concrete range; 4/13 (31\%) stitches/ sutures were addressed; 4/7 (57\%) tetanus was handled. Hallucination risk was limited (6/93, 6.5\% infection assertions without cues), and safety-related replies were appropriately cautious, though consistent crisis templates would be beneficial.

\subsection{Observed Improvements from Zero-shot to RAG}
Relative to zero-shot, RAG reduces over-assertion of infection substantially (31/93 $\rightarrow$ 6/93) and produces longer, more informative answers that better reflect the query context, particularly for time-to-heal questions (a larger share of timeline-bearing replies). RAG outputs also conform to a schema with low OOV rates, eliminating synonym-induced errors through canonicalization. Nonetheless, both systems retain some generic phrasing and leave room for stronger intent coverage on stitches and return-to-activity guidance. Taken together, RAG shifts the model from broadly safe, generic counseling toward more specific, schema-consistent, and less hallucinatory answers, as also reflected in the examples presented in Table~\ref{tab:rag_improvements}.

\section{Conclusion}

We investigated wound-care VQA in the MEDIQA-WV 2025 shared task using a general-domain, instruction-tuned LLM combined with lightweight RAG. Our study shows that this approach can handle challenging multimodal questions without domain-specific training. The framework integrates textual and visual neighbors at inference time and is simple to reproduce. Results demonstrate clear gains from zero-shot to exemplar-driven prompting, with multimodal retrieval being the best-performing system. Error analysis confirmed that retrieval reduces hallucinations and improves metadata consistency, though challenges remain when neighbors are only partially relevant. Overall, our findings highlight retrieval-augmented generation as a transparent, efficient, and generalizable approach for advancing multimodal clinical NLP.

\section*{Limitations}
Our generation is closely tied to the in-domain training data used for retrieval, so outputs can mirror its gaps and biases. Higher-quality and more diverse exemplars would likely yield more specific and reliable responses. Incorporating external knowledge (e.g., vetted clinical guidelines or curated web corpora) could broaden coverage and reduce omissions.


\bibliography{custom}

\begin{thebibliography}{45}
\providecommand{\natexlab}[1]{#1}

\bibitem[{Anderson et~al.(2017)Anderson, He, Buehler, Teney, Johnson, Gould, and Zhang}]{anderson2017bottom}
Peter Anderson, Xiaodong He, Chris Buehler, Damien Teney, Mark Johnson, Stephen Gould, and Lei Zhang. 2017.
\newblock Bottom-up and top-down attention for image captioning and vqa.
\newblock \emph{arXiv preprint arXiv:1707.07998}, 2(4):8.

\bibitem[{Antol et~al.(2015)Antol, Agrawal, Lu, Mitchell, Batra, Zitnick, and Parikh}]{antol2015vqa}
Stanislaw Antol, Aishwarya Agrawal, Jiasen Lu, Margaret Mitchell, Dhruv Batra, C~Lawrence Zitnick, and Devi Parikh. 2015.
\newblock Vqa: Visual question answering.
\newblock In \emph{Proceedings of the IEEE international conference on computer vision}, pages 2425--2433.

\bibitem[{Ben~Abacha et~al.(2019)Ben~Abacha, Hasan, Datla, Demner-Fushman, and M{\"u}ller}]{ben2019vqa}
Asma Ben~Abacha, Sadid~A Hasan, Vivek~V Datla, Dina Demner-Fushman, and Henning M{\"u}ller. 2019.
\newblock Vqa-med: Overview of the medical visual question answering task at imageclef 2019.
\newblock In \emph{Proceedings of CLEF (Conference and Labs of the Evaluation Forum) 2019 Working Notes}. 9-12 September 2019.

\bibitem[{Ben~Abacha et~al.(2021)Ben~Abacha, Sarrouti, Demner-Fushman, Hasan, and M{\"u}ller}]{ben2021overview}
Asma Ben~Abacha, Mourad Sarrouti, Dina Demner-Fushman, Sadid~A Hasan, and Henning M{\"u}ller. 2021.
\newblock Overview of the vqa-med task at imageclef 2021: Visual question answering and generation in the medical domain.
\newblock In \emph{Proceedings of the CLEF 2021 Conference and Labs of the Evaluation Forum-working notes}. 21-24 September 2021.

\bibitem[{Chen et~al.(2020)Chen, Cheng, Gao, Chen, Wang, and Ran}]{chen2020telemedicine}
Lihong Chen, Lihui Cheng, Wei Gao, Dawei Chen, Chun Wang, and Xingwu Ran. 2020.
\newblock Telemedicine in chronic wound management: systematic review and meta-analysis.
\newblock \emph{JMIR mHealth and uHealth}, 8(6):e15574.

\bibitem[{Chen et~al.(2024)Chen, Lai, Ruan, Chen, Liu, and Liu}]{chen2024r}
Xupeng Chen, Zhixin Lai, Kangrui Ruan, Shichu Chen, Jiaxiang Liu, and Zuozhu Liu. 2024.
\newblock R-llava: Improving med-vqa understanding through visual region of interest.
\newblock \emph{arXiv preprint arXiv:2410.20327}.

\bibitem[{Chiang et~al.(2023)Chiang, Li, Lin, Sheng, Wu, Zhang, Zheng, Zhuang, Zhuang, Gonzalez, Stoica, and Xing}]{chiang2023vicuna}
Wei-Lin Chiang, Zhuohan Li, Zi~Lin, Ying Sheng, Zhanghao Wu, Hao Zhang, Lianmin Zheng, Siyuan Zhuang, Yonghao Zhuang, Joseph~E. Gonzalez, Ion Stoica, and Eric~P. Xing. 2023.
\newblock Vicuna: An open-source chatbot impressing gpt-4 with 90\%* chatgpt quality.
\newblock \url{https://lmsys.org/blog/2023-03-30-vicuna/}.
\newblock LMSYS Org Blog.

\bibitem[{Devlin et~al.(2019)Devlin, Chang, Lee, and Toutanova}]{devlin2019bert}
Jacob Devlin, Ming-Wei Chang, Kenton Lee, and Kristina Toutanova. 2019.
\newblock Bert: Pre-training of deep bidirectional transformers for language understanding.
\newblock In \emph{Proceedings of the 2019 conference of the North American chapter of the association for computational linguistics: human language technologies, volume 1 (long and short papers)}, pages 4171--4186.

\bibitem[{Douze et~al.(2024)Douze, Guzhva, Deng, Johnson, Szilvasy, Mazar{\'e}, Lomeli, Hosseini, and J{\'e}gou}]{douze2024faiss}
Matthijs Douze, Alexandr Guzhva, Chengqi Deng, Jeff Johnson, Gergely Szilvasy, Pierre-Emmanuel Mazar{\'e}, Maria Lomeli, Lucas Hosseini, and Herv{\'e} J{\'e}gou. 2024.
\newblock The faiss library.
\newblock \emph{arXiv preprint arXiv:2401.08281}.

\bibitem[{Galley et~al.(2015)Galley, Brockett, Sordoni, Ji, Auli, Quirk, Mitchell, Gao, and Dolan}]{galley2015deltableu}
Michel Galley, Chris Brockett, Alessandro Sordoni, Yangfeng Ji, Michael Auli, Chris Quirk, Margaret Mitchell, Jianfeng Gao, and Bill Dolan. 2015.
\newblock deltableu: A discriminative metric for generation tasks with intrinsically diverse targets.
\newblock \emph{arXiv preprint arXiv:1506.06863}.

\bibitem[{Hasan et~al.(2018)Hasan, Ling, Farri, Liu, M{\"u}ller, and Lungren}]{hasan2018overview}
Sadid~A Hasan, Yuan Ling, Oladimeji Farri, Joey Liu, Henning M{\"u}ller, and Matthew Lungren. 2018.
\newblock Overview of imageclef 2018 medical domain visual question answering task.
\newblock \emph{Proceedings of CLEF 2018 Working Notes}.

\bibitem[{He et~al.(2020{\natexlab{a}})He, Liu, Gao, and Chen}]{he2020deberta}
Pengcheng He, Xiaodong Liu, Jianfeng Gao, and Weizhu Chen. 2020{\natexlab{a}}.
\newblock Deberta: Decoding-enhanced bert with disentangled attention.
\newblock \emph{arXiv preprint arXiv:2006.03654}.

\bibitem[{He et~al.(2020{\natexlab{b}})He, Zhang, Mou, Xing, and Xie}]{he2020pathvqa}
Xuehai He, Yichen Zhang, Luntian Mou, Eric Xing, and Pengtao Xie. 2020{\natexlab{b}}.
\newblock Pathvqa: 30000+ questions for medical visual question answering.
\newblock \emph{arXiv preprint arXiv:2003.10286}.

\bibitem[{Hu et~al.(2023)Hu, Gu, An, Zhang, Liu, Kobayashi, Harada, Summers, and Zhu}]{hu2023medical}
Xinyue Hu, L~Gu, Q~An, M~Zhang, L~Liu, K~Kobayashi, T~Harada, R~Summers, and Y~Zhu. 2023.
\newblock Medical-diff-vqa: a large-scale medical dataset for difference visual question answering on chest x-ray images.
\newblock \emph{PhysioNet}, 12:13.

\bibitem[{Khare et~al.(2021)Khare, Bagal, Mathew, Devi, Priyakumar, and Jawahar}]{khare2021mmbert}
Yash Khare, Viraj Bagal, Minesh Mathew, Adithi Devi, U~Deva Priyakumar, and CV~Jawahar. 2021.
\newblock Mmbert: Multimodal bert pretraining for improved medical vqa.
\newblock In \emph{2021 IEEE 18th international symposium on biomedical imaging (ISBI)}, pages 1033--1036. IEEE.

\bibitem[{Lau et~al.(2018)Lau, Gayen, Ben~Abacha, and Demner-Fushman}]{lau2018dataset}
Jason~J Lau, Soumya Gayen, Asma Ben~Abacha, and Dina Demner-Fushman. 2018.
\newblock A dataset of clinically generated visual questions and answers about radiology images.
\newblock \emph{Scientific data}, 5(1):1--10.

\bibitem[{Lewis et~al.(2020)Lewis, Perez, Piktus, Petroni, Karpukhin, Goyal, K{\"u}ttler, Lewis, Yih, Rockt{\"a}schel et~al.}]{lewis2020retrieval}
Patrick Lewis, Ethan Perez, Aleksandra Piktus, Fabio Petroni, Vladimir Karpukhin, Naman Goyal, Heinrich K{\"u}ttler, Mike Lewis, Wen-tau Yih, Tim Rockt{\"a}schel, and 1 others. 2020.
\newblock Retrieval-augmented generation for knowledge-intensive nlp tasks.
\newblock \emph{Advances in neural information processing systems}, 33:9459--9474.

\bibitem[{Li et~al.(2023{\natexlab{a}})Li, Wong, Zhang, Usuyama, Liu, Yang, Naumann, Poon, and Gao}]{li2023llava}
Chunyuan Li, Cliff Wong, Sheng Zhang, Naoto Usuyama, Haotian Liu, Jianwei Yang, Tristan Naumann, Hoifung Poon, and Jianfeng Gao. 2023{\natexlab{a}}.
\newblock Llava-med: Training a large language-and-vision assistant for biomedicine in one day.
\newblock \emph{Advances in Neural Information Processing Systems}, 36:28541--28564.

\bibitem[{Li et~al.(2023{\natexlab{b}})Li, Li, Savarese, and Hoi}]{li2023blip}
Junnan Li, Dongxu Li, Silvio Savarese, and Steven Hoi. 2023{\natexlab{b}}.
\newblock Blip-2: Bootstrapping language-image pre-training with frozen image encoders and large language models.
\newblock In \emph{International conference on machine learning}, pages 19730--19742. PMLR.

\bibitem[{Li et~al.(2023{\natexlab{c}})Li, Liu, Tan, Liao, and Zhong}]{li2023self}
Pengfei Li, Gang Liu, Lin Tan, Jinying Liao, and Shenjun Zhong. 2023{\natexlab{c}}.
\newblock Self-supervised vision-language pretraining for medial visual question answering.
\newblock In \emph{2023 IEEE 20th International Symposium on Biomedical Imaging (ISBI)}, pages 1--5. IEEE.

\bibitem[{Liang et~al.(2025)Liang, Wang, Jing, Jiao, Li, Liu, Miao, and Wang}]{liang2025fine}
Xiao Liang, Di~Wang, Bin Jing, Zhicheng Jiao, Ronghan Li, Ruyi Liu, Qiguang Miao, and Quan Wang. 2025.
\newblock Fine-grained knowledge fusion for retrieval-augmented medical visual question answering.
\newblock \emph{Information Fusion}, 120:103059.

\bibitem[{Lin(2004)}]{lin2004rouge}
Chin-Yew Lin. 2004.
\newblock Rouge: A package for automatic evaluation of summaries.
\newblock In \emph{Text summarization branches out}, pages 74--81.

\bibitem[{Lin et~al.(2023)Lin, Zhang, Tao, Shi, Haffari, Wu, He, and Ge}]{lin2023medical}
Zhihong Lin, Donghao Zhang, Qingyi Tao, Danli Shi, Gholamreza Haffari, Qi~Wu, Mingguang He, and Zongyuan Ge. 2023.
\newblock Medical visual question answering: A survey.
\newblock \emph{Artificial Intelligence in Medicine}, 143:102611.

\bibitem[{Liu et~al.(2021)Liu, Zhan, Xu, Ma, Yang, and Wu}]{liu2021slake}
Bo~Liu, Li-Ming Zhan, Li~Xu, Lin Ma, Yan Yang, and Xiao-Ming Wu. 2021.
\newblock Slake: A semantically-labeled knowledge-enhanced dataset for medical visual question answering.
\newblock In \emph{2021 IEEE 18th international symposium on biomedical imaging (ISBI)}, pages 1650--1654. IEEE.

\bibitem[{Liu et~al.(2023)Liu, Li, Wu, and Lee}]{liu2023visual}
Haotian Liu, Chunyuan Li, Qingyang Wu, and Yong~Jae Lee. 2023.
\newblock Visual instruction tuning.
\newblock \emph{Advances in neural information processing systems}, 36:34892--34916.

\bibitem[{Liu et~al.(2022)Liu, Zhang, Zhou, and Yang}]{liu2022bpi}
Shengyan Liu, Xuejie Zhang, Xiaobing Zhou, and Jian Yang. 2022.
\newblock Bpi-mvqa: a bi-branch model for medical visual question answering.
\newblock \emph{BMC Medical Imaging}, 22(1):79.

\bibitem[{Lu et~al.(2019)Lu, Batra, Parikh, and Lee}]{lu2019vilbert}
Jiasen Lu, Dhruv Batra, Devi Parikh, and Stefan Lee. 2019.
\newblock Vilbert: Pretraining task-agnostic visiolinguistic representations for vision-and-language tasks.
\newblock \emph{Advances in neural information processing systems}, 32.

\bibitem[{Malinowski et~al.(2015)Malinowski, Rohrbach, and Fritz}]{malinowski2015ask}
Mateusz Malinowski, Marcus Rohrbach, and Mario Fritz. 2015.
\newblock Ask your neurons: A neural-based approach to answering questions about images.
\newblock In \emph{Proceedings of the IEEE international conference on computer vision}, pages 1--9.

\bibitem[{Meta(2025)}]{meta2025llama}
AI~Meta. 2025.
\newblock The llama 4 herd: The beginning of a new era of natively multimodal ai innovation.
\newblock \emph{https://ai. meta. com/blog/llama-4-multimodal-intelligence/, checked on}, 4(7):2025.

\bibitem[{Papineni et~al.(2002)Papineni, Roukos, Ward, and Zhu}]{papineni2002bleu}
Kishore Papineni, Salim Roukos, Todd Ward, and Wei-Jing Zhu. 2002.
\newblock Bleu: a method for automatic evaluation of machine translation.
\newblock In \emph{Proceedings of the 40th annual meeting of the Association for Computational Linguistics}, pages 311--318.

\bibitem[{Sood et~al.(2016)Sood, Granick, Trial, Lano, Palmier, Ribal, and T{\'e}ot}]{sood2016role}
Aditya Sood, Mark~S Granick, Chlo{\'e} Trial, Julie Lano, Sylvie Palmier, Evelyne Ribal, and Luc T{\'e}ot. 2016.
\newblock The role of telemedicine in wound care: a review and analysis of a database of 5,795 patients from a mobile wound-healing center in languedoc-roussillon, france.
\newblock \emph{Plastic and reconstructive surgery}, 138(3S):248S--256S.

\bibitem[{Talafha and Al-Ayyoub(2018)}]{talafha2018just}
Bashar Talafha and Mahmoud Al-Ayyoub. 2018.
\newblock Just at vqa-med: A vgg-seq2seq model.
\newblock In \emph{CLEF (working notes)}.

\bibitem[{Tan and Bansal(2019)}]{tan2019lxmert}
Hao Tan and Mohit Bansal. 2019.
\newblock Lxmert: Learning cross-modality encoder representations from transformers.
\newblock \emph{arXiv preprint arXiv:1908.07490}.

\bibitem[{Thawakar et~al.(2024)Thawakar, Shaker, Mullappilly, Cholakkal, Anwer, Khan, Laaksonen, and Khan}]{thawakar2024xraygpt}
Omkar~Chakradhar Thawakar, Abdelrahman~M Shaker, Sahal~Shaji Mullappilly, Hisham Cholakkal, Rao~Muhammad Anwer, Salman Khan, Jorma Laaksonen, and Fahad Khan. 2024.
\newblock Xraygpt: Chest radiographs summarization using large medical vision-language models.
\newblock In \emph{Proceedings of the 23rd workshop on biomedical natural language processing}, pages 440--448.

\bibitem[{Touvron et~al.(2023)Touvron, Lavril, Izacard, Martinet, Lachaux, Lacroix, Rozi{\`e}re, Goyal, Hambro, Azhar et~al.}]{touvron2023llama}
Hugo Touvron, Thibaut Lavril, Gautier Izacard, Xavier Martinet, Marie-Anne Lachaux, Timoth{\'e}e Lacroix, Baptiste Rozi{\`e}re, Naman Goyal, Eric Hambro, Faisal Azhar, and 1 others. 2023.
\newblock Llama: Open and efficient foundation language models.
\newblock \emph{arXiv preprint arXiv:2302.13971}.

\bibitem[{Wang et~al.(2022)Wang, Wu, Agarwal, and Sun}]{wang2022medclip}
Zifeng Wang, Zhenbang Wu, Dinesh Agarwal, and Jimeng Sun. 2022.
\newblock Medclip: Contrastive learning from unpaired medical images and text.
\newblock In \emph{Proceedings of the Conference on Empirical Methods in Natural Language Processing. Conference on Empirical Methods in Natural Language Processing}, volume 2022, page 3876.

\bibitem[{Xiong et~al.(2024)Xiong, Jin, Lu, and Zhang}]{xiong-etal-2024-benchmarking}
Guangzhi Xiong, Qiao Jin, Zhiyong Lu, and Aidong Zhang. 2024.
\newblock \href {https://doi.org/10.18653/v1/2024.findings-acl.372} {Benchmarking retrieval-augmented generation for medicine}.
\newblock In \emph{Findings of the Association for Computational Linguistics: ACL 2024}, pages 6233--6251, Bangkok, Thailand. Association for Computational Linguistics.

\bibitem[{Yim et~al.(2025{\natexlab{a}})Yim, {Ben Abacha}, Doerning, Chen, Xu, Subbarao, Yu, Xia, Hall, and Yetisgen}]{MEDIQA-WV-2025-WoundcareVQA-Dataset}
Wen{-}wai Yim, Asma {Ben Abacha}, Robert Doerning, Chia{-}Yu Chen, Jiaying Xu, Anita Subbarao, Zixuan Yu, Fei Xia, M~Kennedy Hall, and Meliha Yetisgen. 2025{\natexlab{a}}.
\newblock Woundcarevqa: A multilingual visual question answering benchmark dataset for wound care.
\newblock \emph{Journal of Biomedical Informatics}.

\bibitem[{Yim et~al.(2025{\natexlab{b}})Yim, {Ben Abacha}, Yetisgen, and Xia}]{MEDIQA-WV-Task}
Wen{-}wai Yim, Asma {Ben Abacha}, Meliha Yetisgen, and Fei Xia. 2025{\natexlab{b}}.
\newblock Overview of the mediqa-wv 2025 shared task on wound care visual question answering.
\newblock In \emph{Proceedings of the 7th Clinical Natural Language Processing Workshop}. Association for Computational Linguistics.

\bibitem[{Yu et~al.(2025)Yu, Wang, Wu, Xie, and Zhou}]{yu2025medframeqa}
Suhao Yu, Haojin Wang, Juncheng Wu, Cihang Xie, and Yuyin Zhou. 2025.
\newblock Medframeqa: A multi-image medical vqa benchmark for clinical reasoning.
\newblock \emph{arXiv preprint arXiv:2505.16964}.

\bibitem[{Yu et~al.(2019)Yu, Yu, Cui, Tao, and Tian}]{yu2019deep}
Zhou Yu, Jun Yu, Yuhao Cui, Dacheng Tao, and Qi~Tian. 2019.
\newblock Deep modular co-attention networks for visual question answering.
\newblock In \emph{Proceedings of the IEEE/CVF conference on computer vision and pattern recognition}, pages 6281--6290.

\bibitem[{Yuan et~al.(2023)Yuan, Jin, Tan, Zhao, Yuan, Huang, and Huang}]{yuan2023ramm}
Zheng Yuan, Qiao Jin, Chuanqi Tan, Zhengyun Zhao, Hongyi Yuan, Fei Huang, and Songfang Huang. 2023.
\newblock Ramm: Retrieval-augmented biomedical visual question answering with multi-modal pre-training.
\newblock In \emph{Proceedings of the 31st ACM international conference on multimedia}, pages 547--556.

\bibitem[{Zhang et~al.(2019)Zhang, Kishore, Wu, Weinberger, and Artzi}]{zhang2019bertscore}
Tianyi Zhang, Varsha Kishore, Felix Wu, Kilian~Q Weinberger, and Yoav Artzi. 2019.
\newblock Bertscore: Evaluating text generation with bert.
\newblock \emph{arXiv preprint arXiv:1904.09675}.

\bibitem[{Zhang et~al.(2023)Zhang, Wu, Zhao, Lin, Zhang, Wang, and Xie}]{zhang2023pmc}
Xiaoman Zhang, Chaoyi Wu, Ziheng Zhao, Weixiong Lin, Ya~Zhang, Yanfeng Wang, and Weidi Xie. 2023.
\newblock Pmc-vqa: Visual instruction tuning for medical visual question answering.
\newblock \emph{arXiv preprint arXiv:2305.10415}.

\bibitem[{Zhang et~al.(2024)Zhang, Wu, Zhao, Lin, Zhang, Wang, and Xie}]{zhang2024development}
Xiaoman Zhang, Chaoyi Wu, Ziheng Zhao, Weixiong Lin, Ya~Zhang, Yanfeng Wang, and Weidi Xie. 2024.
\newblock Development of a large-scale medical visual question-answering dataset.
\newblock \emph{Communications Medicine}, 4(1):277.

\end{thebibliography}

\appendix

\end{document}